# BACON: Deep-Learning Powered AI for Poetry Generation with Author Linguistic Style Transfer

**Alejandro Rodriguez Pascual**
**Campolindo High School**
300 Moraga Rd, Moraga, CA 94556
alejandro.rodriguez19@auhsdschools.org

## Abstract

This paper describes BACON[1], a basic prototype of an automatic poetry generator with author linguistic style transfer. It combines concepts and techniques from finite state machinery, probabilistic models, artificial neural networks and deep learning, to write original poetry with rich aesthetic-qualities in the style of any given author. Extrinsic evaluation of the output generated by BACON shows that participants were unable to tell the difference between human and AI-generated poems in any statistically significant way.

## 1   Introduction

### 1.1   Rationale

Can machines create art? Common notions of creativity and art are often bound up in the idea of a creative process that is special, hard to define, and maybe even magical, and exclusively reserved to the human mind. Machines, and computers in particular, on the other hand, are usually considered systems that function in a fixed, rule-bound manner. The idea that machines or computers could be creative seems far-fetched. But is such a view justified?

Artificial Intelligence (AI) is commonly defined as intelligence displayed by machines, in contrast with the "natural intelligence'" (NI) displayed by humans and other animals. Colloquially, the term "artificial intelligence" is applied when a machine mimics cognitive functions associated with human minds, such as learning and problem solving. Recent advances in computer science theory and technologies, such as machine learning, artificial neural networks, and deep learning have brought AI into the mainstream for a large body of applications, across areas such as computer vision, object detection and recognition, self-driving cars, medical diagnostics, real-time traffic routing, fraud prevention and credit decisions, and many more. Natural Language Processing (NLP) is also an active area of research and development, with advances and applications in automatic language generation and translation, voice recognition, voice-to-text generation, and conversational interfaces (such as AI-powered bots).

These advances poise the question: is art a realm still exclusively reserved to the human mind? Some AI research projects have started to explore this question. Two relevant examples are Shelley (http://shelley.ai) -a project by MIT Media Lab that has developed a deep learning powered AI that generates, in a collaborative way with humans, horror stories very much like those originally written by Mary Shelley-, and Magenta (https://magenta.tensorflow.org) -a Google research project that intends to "advance the state-of-the art in music, video, image and text generation", and to answer the questions: "Can we use machine learning to create compelling art and music? If so, how? If not, why not?"

The question that this project explores is: can a computer generate poetry that resembles that created by any given author, and therefore compose poems that, for all effects, could be attributable to the original author?

---

[1] BACON stands for **B**asic **AI** for **C**ollaborative p**O**etry writi**N**g. The name is coined after Sir Francis Bacon who, according to some, was who actually wrote William Shakespeare's works.

## 1.2 Project Goals

The goal of this project is to develop a software system that automatically generates poetry (in English) with the "flavor"' or style of any given author –what is known as linguistic style transfer. The generated output should consist of original works that are, ideally, of a quality and resemblance high enough as to be indistinguishable from the existing works of such author.

Additional goals include:

1. To explore the most relevant concepts in artificial intelligence that can be applied to human traits such as creativity and artistic creation.
2. To understand and hopefully advance the state-of-the-art in semantic modeling and automatic generation of poetry, as an example of feature-rich, hard to mimic by machines, human creation.
3. To explore how linguistic style transfer can enable more intimate human-AI interactions, allowing us to bridge the inherent gap between AI agents (such as bots) and human beings.

## 1.3 Previous work

Although there is an extensive research body and commercial applications of AI for NLP, and in particular for automatic text generation, most of them fall in the area of prose generation (Perera and Nand, 2017) or machine translation (Bahdanau, Cho, and Bengio, 2014; Luong, Pham, and Manning, 2015). Poetry generation, on the other hand, is a less explored field. Automatic poetry generation is a challenging task, since, in addition to meaning, correct and meaningful poetry must satisfy various form constraints and aesthetic requirements. Some early attempts, dating back to the 1950s, approached the problem with template- and grammar-based generators –see (Oliveira, 2009), (Oliveira, 2012)-, or form-aware generators –see Manurung (2003). More recently, deep learning and artificial neural networks have entered the field, with de Boer (2017), Xie, Rastogi, and Chang (2017), which focus on Shakespeare sonnets, or Yang et al. (2017); Zhang and Lapata, (2014) for Chinese poetry, as relevant examples.

The problem of how to automatically generate poetry with the specific style of a given author can be split into the following three sub-problems:

1. *Linguistic style modeling*: how to model or capture the linguistic style of a given author.
2. *Linguistic style transfer*: how to use that model to guide the automatic generation of poems.
3. *Automatic poem generation*: how to generate meaningful poetry with rich aesthetic rules.

The application of artificial neural networks to the problem of style transfer -dubbed "neural style transfer"'- has been explored and implemented successfully for paintings by Gatys, Ecker, and Bethge (2015). Linguistic style transfer for prose works has also been recently explored by some authors, for example in Ficler and Goldberg (2017), Xu et al. (2012).

BACON builds on the research developed by Jack Hopkins and Douwe Kiela (Hopkins and Kiela, 2017). Their research approaches automatic poetry generation by explicitly splitting the problem into two sub-tasks:

1. The problem of content, concerned with a poem's semantics.
2. The problem of form, concerned with the aesthetic rules that a poem must follow.

Hopkins and Kiela (2017) developed an automatic poetry generator (APG) that combines, in a pipeline, the following two components:

1. A generative language model representing content,
2. A discriminative model representing form.

This approach allows to represent the problem of creating correct and meaningful poetry as "a constraint satisfaction problem imposed on the output of a generative model, where the constraints to restrict the types of generated poetry can be modified at will" (Hopkins and Kiela, 2017).

Accordingly, in linguistic style transfer the modeling of a given author style could consider a set of features that relate to either content or form; these features should be extracted from each poem in the corpus generated by compiling all the author's works. The following is a tentative list of the features that can be considered:

*Content*:
a) The poem's topic, subject matter, and theme.
b) The poem's tone - tone is the poet's attitude toward the subject. It could be positive or negative, joyful, sarcastic, nostalgic, or any other emotion.
c) The poem's word choices and order.
d) The poem's figurative language, such as comparisons (metaphors and similes), word play, alliteration, assonance, consonance, deliberate exaggeration, symbolism, imagery, etc.

*Form*:
a) The poem's structure: generally speaking, structure has to do with the overall organization of lines into stanzas, strophes, verses, etc.
b) The poem's rhythm: the systematic regularity in rhythm in a poem is known as meter, and is usually identified by examining the type of "foot" and the number of "feet".
c) The poem's rhyme: rhyme is the repetition of similar sounds. In poetry, the most common kind of rhyme is the end rhyme, which occurs at the end of two or more lines. Internal rhyme can also be used.

Note that many modern poems may not have any identifiable structure nor follow strict rules of meter or rhyme, especially throughout an entire poem (i.e. they are free verse).

## 2  Proposed approach

BACON approaches the problem of automatic poetry generation with linguistic style transfer by splitting the solution in two components: (1) a linguistic style modeler (LSM), which builds a probabilistic model of the style used by any given author, and, (2) a deep-learning powered automatic poem generator (APG) which uses the model generated by the LSM to guide the generation of original, meaningful poetry, with rich aesthetic rules, in the style of such author. This is illustrated in Figure 1.

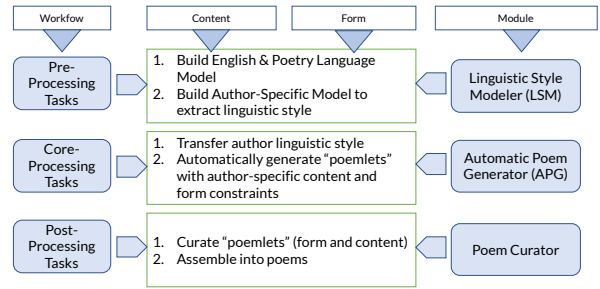

Figure 1 - BACON approach

BACON implements the following features:

1. *Content representation*: BACON uses the concept of Vector Space Models (VSM) for representing words and word sequences (n-grams). Three VSM are constructed from:
   - A large set of English poetry texts, consisting of 7.6 million words and 34.4 million characters, taken from $20^{th}$ century poetry books.
   - A large set of general English language texts, consisting of a full English Wikipedia dump, 5.5 million documents, 43.6 million pages.
   - A given set of author-specific texts (for example, for one such authors, Robinson Jeffers, consisting of 144 works, 135 thousand words).

   A vector space model is an algebraic model for representing text documents (or any objects, in general) as vectors of identifiers, such as, for example, words (or, more generally, n-grams, a contiguous sequence of n items from a given sample of text or speech). A VSM represents (or "embeds") words in a continuous multi-dimensional vector space where semantically similar words are mapped to nearby points (or equivalently, they are "embedded nearby each other"). This is a consequence of the Distributional Hypothesis, which states that words that appear in the same contexts share semantic meaning.

2. *Linguistic style modeling*: BACON provides style modeling by building probabilistic models for feature extraction of the following linguistic style elements: word choices and word order, and latent topics and themes. It uses the concepts of *Term-Frequency-Inverse Document Frequency (TF-IDF)* and *Latent Dirichlet Analysis (LDA)* to this extent. For word choice and word order BACON computes a TF-IDF model of an author's corpus and extracts *high-entropy terms* by selecting the *n*% (configurable parameter) of the terms in the corpus with the highest TF-IDF score. These are defined as those with a high TF-IDF score, computed against the larger corpus of English works. For topic modeling BACON computes an LDA model from an author's corpus and extracts relevant topic/theme words by selecting the *m* (configurable parameter) more relevant topics from the model.

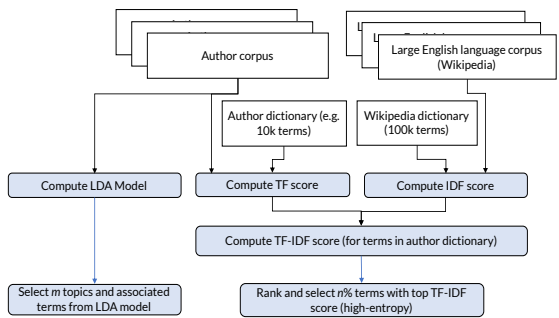

Figure 2 - TF-IDF and LDA models

TF-IDF is a term-weighting scheme applied by search engines, recommender systems, and user modeling engines. TF-IDF was introduced by Jones (1972) and contains two components: term frequency (TF) and inverse document frequency (IDF). TF is the frequency with which a term occurs in a document or user model. The rationale is that the more frequently a term occurs, the more likely this term describes a document's content or user's information need. IDF reflects the relative importance of the term by computing the inverse frequency of documents containing the term within the entire corpus of documents under consideration (for search or recommendation). The basic assumption is that a term should be given a higher weight if few other documents also contain that term, because rare terms will likely be more representative of a document's content or user's interests (Beel, Langer, and Gipp, 2017).

LDA is an unsupervised generative probabilistic model of a corpus. LDA was first introduced by Blei et al. (2003) and is one of the most popular methods in topic modeling. In LDA each item of a collection is modeled as a finite mixture over an underlying set of topics. Each topic is, in turn, modeled as an infinite mixture over an underlying set of topic probabilities. In the context of text modeling, the topic probabilities provide an explicit representation of a document.

3. *Linguistic style transfer* (of content elements) is achieved by probabilistic conditioning/boosting of the high-entropy n-grams and topic words applied to the Automatic Poetry Generation (APG) module.

4. *Automatic Poetry Generation (APG):* the APG module is implemented as a pipeline of (1) a content generator, followed by (2) a form shaper. The content generator uses a character-level generative model representing content, implemented by a *Long Short-Term Memory (LSTM) Recurrent Neural Network (RNN)*. The form shaper uses a *Weighted Finite State Transducer (WFST)-* based discriminative model representing form. This APG is a modified version of the one built by Hopkins and Kiela (2017)[2].

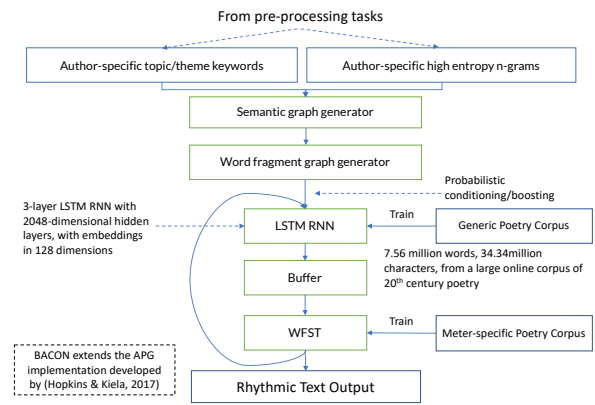

Figure 3 - Automatic Poem Generator (APG)

Recurrent Neural Networks (RNN) are artificial neural networks with recurrent connections, allowing them to learn temporal regularities and model sequences. Long Short-Term Memory (LSTM) (Hochreiter and Schmidhuber, 1997) is a recurrent neural network architecture that overcomes the problem of exponentially vanishing of gradients, allowing RNNs to be trained many time-steps into the past, and therefore to learn more complex programs (Schmidhuber, 2015). LSTMs and related architectures have been proved successful in sequence generation in several domains such as music (Boulanger-Lewandowski, Bengio, and Vincent, 2012; Nayebi and Vitelli, 2015), text (Sutskever, Martens, and Hinton, 2011), images (Gregor et al., 2015), machine translation (Sutskever, Vinyals, and Le, 2014), and speech synthesis (Wu and King, 2016).

---
[2] A pre-trained model was provided by the authors of Hopkins and Kiela (2017).

A finite-state transducer (FST) is a type of finite-state automaton whose state transitions are labeled with both input and output symbols, and therefore it maps between two sets of symbols. A path through the transducer encodes a mapping from an input symbol sequence to an output symbol sequence. A weighted transducer adds weights on transitions to the input and output symbols of a FST (Vidal et al., 2005). These weights may encode any quantity that accumulates along paths (e.g. probabilities or durations) and allow to compute the overall weight of mapping an input sequence to an output sequence. Weighted transducers are thus a natural choice to represent probabilistic finite-state models, which are commonly used for pattern recognition, or in fields to which pattern recognition is linked, such as computational linguistics, machine learning, time series analysis, computational biology, speech recognition, and machine translation.

## 3 Workflow

The full workflow of BACON can be split into pre-processing tasks, core-processing tasks, and post-processing tasks, as illustrated in Figure 4 and Figure 5.

1. Pre-processing tasks: all the tasks required to capture the style of a given author.
2. Core-processing tasks: tasks required to automatically generate poems (to be more precise, in its current form, BACON generates "poemlets", or small poem fragments).
3. Post-processing tasks: task required to curate and assemble larger poems.

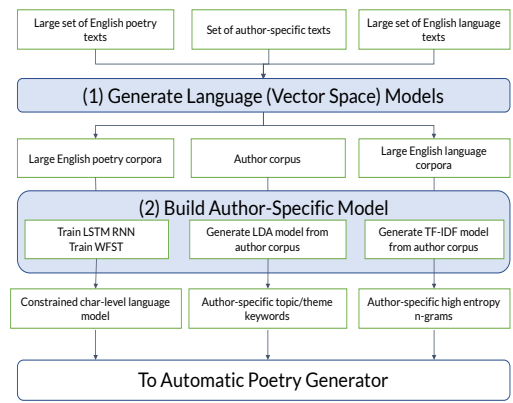

Figure 4 - Workflow: pre-processing tasks

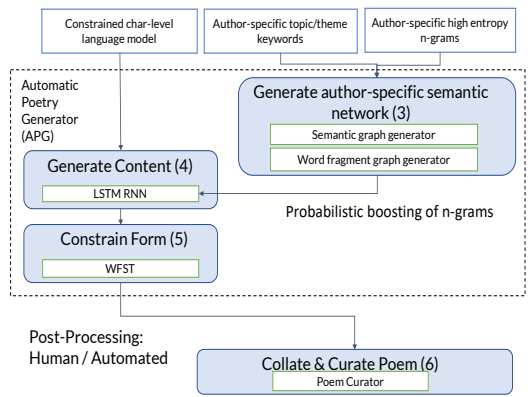

Figure 5 - Workflow: core and post-processing tasks

## 4 Software architecture

BACON combines a set of open source software modules with software that has been developed specifically for this project. Figure 6 illustrates the various software modules and libraries used in BACON.

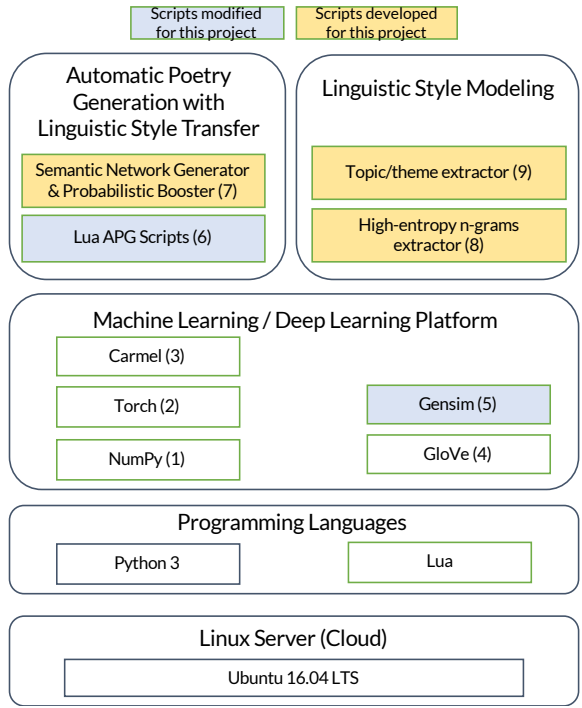

Figure 6 - Software platform

Notes:
(1) The fundamental package for scientific computing with Python, licensed under the BSD license.
(2) An open-source Deep Learning (DL) platform initially developed by Facebook AI Research.
(3) A WFST package written by Jonathan Graehl at USC/ISI used to create pronunciation models, licensed under the Apache License, v2.0.
(4) An unsupervised learning algorithm for obtaining vector representations for words, developed by Stanford's NLP research group, licensed under the Apache License, v2.0.
(5) A Python package to perform unsupervised semantic modelling from plain text, developed by Rehurek and Sojka (2010), licensed under the OSI-approved GNU LGPLv2.1 license.
(6) Lua scripts for automatic poetry generation, developed by Hopkins and Kiela (2017).
(7) Lua scripts for semantic network generation and probabilistic boosting.
(8) and (9) Python scripts for TF-IDF and LDA modeling.

## 5 Output

So far, BACON has been tested on two authors: Robinson Jeffers, and John Keats. We show here the output generated by linguistic style transfer on Robinson Jeffers, an American poet, known for his work about the central California coast, and considered by many as a formidable presence in modern literature. "Jefferson Robins" is a fictional author, an incarnation of BACON that takes as input the poetry works of Jefferson Robins. The corpus for this author was generated by processing 144 of his poetry works, for a total of 135 thousand words. BACON has been used to generate 12 poems "written by Jefferson Robins".

A full list of generated poems can be found in the project GitHub site
(https://github.com/alexrodpas/BACON)

---

**Praise Life – by Robinson Jeffers**

This country least, but every inhabited country
Is clotted with human anguish.
Remember that at your feasts.
And this is no new thing but from time out of mind,
No transient thing, but exactly
Conterminous with human life.
Praise life, it deserves praise, but the praise of life
That forgets the pain is a pebble
Rattled in a dry gourd.

---

**The Mustard-Bloom of Gold – by "Jefferson Robins"**

In the trials of extraordinary war
a week ago, a stranger came along,
Robert Johnson from France.
He told about his loss and death,
his broken bones, the sound of the stars,
the children that buy the books
and paint on them every day.
He said, "the wind began to burn
and the iron sky to fall
that same day, when the sun waited for me
to breathe the mustard-bloom of gold and stripes".

---

Interestingly, "Jefferson Robins" seems to have some interests of its own, since an LDA model of its generated works (although its literary works are still just a few) shows some topics not present in the LDA model computed for Robinson Jeffers.

## 6 Project evaluation

BACON was evaluated by an extrinsic evaluation procedure, performed by conducting an indistinguishability study with a selection of poems written by a human poet and automatically generated poems in the style of that same author -a variant of the so-called Turing test for art works. Boden (2010) defines the criterion as follows: for a program to pass the Turing Test for art works 'would be for it to produce artwork which was (1) indistinguishable from one produced by a human being and/or (2) was seen as having as much aesthetic value as one produced by a human being'.

To this extent, 62 human participants - high school students, and adults (teachers and staff, parents, any other adult)- were used for evaluation of the output generated by BACON. Participants, who were not financially incentivized, thus perceiving the evaluation as a purely intellectual challenge, were asked to fill out an online survey, in which they were presented with a few poems at random, and asked:

a) About each poem's aesthetic and poetic qualities: participants were asked to rate each poem on Readability, Evocativeness and Aesthetics.
b) Whether they thought that each poem was composed either by a given human author, or by a software system.

These results show that:

1. The poems generated by BACON got scores close to the original ones in all 3 measures.
2. The failure ratio in author identification was very similar between original poems and those generated by BACON (0.41 vs. 0.48 respectively).
3. Participants were unable to tell the difference between human and AI-generated poems in any statistically significant way (Chi$^2$ test, p=0.85).

An intrinsic evaluation of BACON will be performed in the future. This will be done by using automatic evaluation measures similar to BLEU (Papineni et al., 2002), an evaluation methodology which compares the content of a generated text against human-written gold standards using word or n-gram overlap.

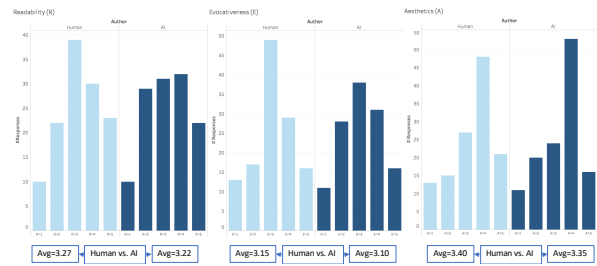

Figure 7 - Evaluation of aesthetic qualities of output

## 7 Conclusions and future work

Extrinsic evaluation of the output generated by BACON (for example in the style of Robinson Jeffers) shows that this approach to linguistic style transfer is promising. TF-IDF modeling and LDA modeling, combined with probabilistic boosting of high-entropy n-grams in the APG provide a basic form of linguistic style transfer.

However, the generated output is still somewhat crude and needs some human post-processing for content curation and composition of full-length poems. Additional work will be necessary to improve the syntactic quality of generated output, its semantic coherence, and fine tune linguistic style modeling and transfer.

Regarding future work, the following topics will be explored:

1. Develop probabilistic models that capture elements of poem's form for linguistic style transfer.
2. Enhance topic modeling for small-size corpora to better model author style. Miao, Grefenstette, and Blunsom (2017) shows some promising approaches.
3. Enhance content modeling for linguistic style transfer. Explore the possible application of concepts such as skip-thought vectors (Kiros et al., 2015).
4. Explore and evaluate alternative approaches to linguistic style transfer. Some researchers are exploring alternatives to

probabilistic models for style transfer. In particular, both Carlson, Riddell, and Rockmore (2017) and Kabbara and Cheung (2016) apply RNN to related problems of style transfer in machine translation and natural language generation.

5. Develop scripts to automate post-processing tasks in the workflow.
6. Better integrate pre-, core, and post-processing scripts into a more continuous workflow. Parametrize code for pre- and post-processing tasks.
7. Improve accuracy of the RNN and WFST used in the APG. Use GPU-based computational resources.
8. Make the full code more future-proof. Migrate all code to PyTorch or TensorFlow.
9. Explore the following interesting variation: "author hybridization" –mixing corpora of two different authors into one, e.g. Shakespeare and Emily Dickinson.

## Acknowledgments

I am grateful to Jack Hopkins (Cambridge University, UK) and Douwe Kiela (Facebook AI Research) for their initial guidance and for providing me with a pre-trained model for the LSTM RNN and the WFST used in the APG module.

## References

Bahdanau, D., Cho, K., Bengio, Y. (2014). Neural Machine Translation by Jointly Learning to Align and Translate. *ArXiv:1409.0473 [Cs, Stat]*. Retrieved from http://arxiv.org/abs/1409.0473

Beel, J., Langer, S., Gipp, B. (2017). TF-IDuF: A Novel Term-Weighting Scheme for User Modeling based on Users' Personal Document Collections. *Proceedings of the 12th IConference. March 2017.*

Blei, D., Ng, A., and Jordan, M. (2003). Latent Dirichlet allocation. *Journal of Machine Learning Research, 3:993–1022.*

Boden, M. A. (2010). The Turing test and artistic creativity. *Kybernetes*, *39*(3), 409–413. https://doi.org/10.1108/03684921011036132

Boulanger-Lewandowski, N., Bengio, Y., Vincent, P. (2012). Modeling Temporal Dependencies in High-Dimensional Sequences: Application to Polyphonic Music Generation and Transcription. *ArXiv:1206.6392 [Cs, Stat]*. Retrieved from http://arxiv.org/abs/1206.6392

Carlson, K., Riddell, A., Rockmore, D. (2017). Zero-Shot Style Transfer in Text Using Recurrent Neural Networks. *ArXiv Preprint ArXiv:1711.04731*.

de Boer, T. S. (2017). Syllable and Meter in Neural Network generated Shakespeare-like Sonnets.

Ficler, J., Goldberg, Y. (2017). Controlling linguistic style aspects in neural language generation. *ArXiv Preprint ArXiv:1707.02633*.

Gatys, L. A., Ecker, A. S., Bethge, M. (2015). A Neural Algorithm of Artistic Style. *ArXiv:1508.06576 [Cs,q-Bio]*. Retrieved from http://arxiv.org/abs/1508.06576

Gregor, K., Danihelka, I., Graves, A., Rezende, D. J., Wierstra, D. (2015). DRAW: A Recurrent Neural Network For Image Generation. *ArXiv:1502.04623 [Cs]*. Retrieved from http://arxiv.org/abs/1502.04623

Hochreiter, S., Schmidhuber, J. (1997). Long Short-Term Memory. *Neural Comput.*, *9*(8), 1735–1780. https://doi.org/10.1162/neco.1997.9.8.1735

Hopkins, J., Kiela, D. (2017). Automatically Generating Rhythmic Verse with Neural Networks. In *Proceedings of the 55th Annual Meeting of the Association for Computational Linguistics (Volume 1: Long Papers)* (Vol. 1, pp. 168–178).

Jones, K. S. (1972). A statistical interpretation of term specificity and its application in retrieval. *Journal of Documentation*, *28*(1), 11–21.

Kabbara, J., Cheung, J. C. K. (2016). Stylistic Transfer in Natural Language Generation Systems Using Recurrent Neural Networks. *EMNLP 2016*, 43.

Kiros, R., Zhu, Y., Salakhutdinov, R. R., Zemel, R., Urtasun, R., Torralba, A., Fidler, S. (2015). Skip-thought vectors. In *Advances in neural information processing systems* (pp. 3294–3302).

Luong, M.-T., Pham, H., Manning, C. D. (2015). Effective Approaches to Attention-based Neural Machine Translation. *ArXiv:1508.04025 [Cs]*. Retrieved from http://arxiv.org/abs/1508.04025

Manurung, H. M. (2003). An Evolutionary Algorithm Approach to Poetry Generation.

Miao, Y., Grefenstette, E., Blunsom, P. (2017). Discovering Discrete Latent Topics with Neural Variational Inference. *ArXiv:1706.00359 [Cs]*. Retrieved from http://arxiv.org/abs/1706.00359

Nayebi, A., Vitelli, M. (2015). Gruv: Algorithmic music generation using recurrent neural networks. *Course CS224D: Deep Learning for Natural Language Processing (Stanford)*.

Oliveira, H. (2009). Automatic generation of poetry: an overview. Universidade de Coimbra.

Oliveira, H. (2012). PoeTryMe: Towards Meaningful Poetry Generation.

Papineni, K., Roukos, S., Ward, T., Zhu, W.-J. (2002). BLEU: a method for automatic evaluation of machine translation. *In Proceedings of the 40th annual meeting on association for computational linguistics (pp. 311–318). Association for Computational Linguistics.*

Perera, R., Nand, P. (2017). Recent Advances in Natural Language Generation: A Survey and Classification of the Empirical Literature. *Computing and Informatics, 36(1), 1–32.*

Software framework for topic modelling with large corpora. Řehůřek, R., Sojka, P. (2010). *Proc. LREC Workshop on New Challenges for NLP Frameworks.*

Schmidhuber, J. (2015). Deep Learning in Neural Networks: An Overview. *Neural Networks, 61, 85–117.* https://doi.org/10.1016/j.neunet.2014.09.003

Sutskever, I., Martens, J., Hinton, G. E. (2011). Generating text with recurrent neural networks. *In Proceedings of the 28th International Conference on Machine Learning (ICML-11) (pp. 1017–1024).*

Sutskever, I., Vinyals, O., Le, Q. V. (2014). Sequence to sequence learning with neural networks. *In Advances in neural information processing systems (pp. 3104–3112).*

Vidal, E., Thollard, F., Higuera, C. de la, Casacuberta, F., Carrasco, R. C. (2005). Probabilistic finite- state machines - part I. *IEEE Transactions on Pattern Analysis and Machine Intelligence, 27(7), 1013–1025.* https://doi.org/10.1109/TPAMI.2005.147

Wu, Z., King, S. (2016). Investigating gated recurrent neural networks for speech synthesis. *ArXiv:1601.02539 [Cs].* Retrieved from http://arxiv.org/abs/1601.02539

Xie, S., Rastogi, R., Chang, M. (2017). Deep Poetry: Word-Level and Character-Level Language Models for Shakespearean Sonnet Generation.

Xu, W., Ritter, A., Dolan, B., Grishman, R., Cherry, C. (2012). Paraphrasing for style. *Proceedings of COLING 2012, 2899–2914.*

Yang, X., Lin, X., Suo, S., Li, M. (2017). Generating Thematic Chinese Poetry with Conditional Variational Autoencoder.

Zhang, X., Lapata, M. (2014). Chinese Poetry Generation with Recurrent Neural Networks. https://doi.org/10.3115/v1/D14-1074